%% file: main.tex
\documentclass{article}
\usepackage{spconf,amsmath,graphicx}

\usepackage{amsfonts}
\usepackage{url}
\usepackage{multicol}
\usepackage{multirow}
\usepackage{hyperref}
\usepackage{cleveref}
\usepackage[font=small,skip=0pt]{caption}
\usepackage{subfig}

\title{Automatic Severity Classification of Dysarthric speech \\ by using Self-supervised Model with Multi-task Learning}

\name{
    Eun Jung Yeo\textsuperscript{1}$^*$\thanks{$^*$Equal contributors.},
    Kwanghee Choi\textsuperscript{2}$^*$,
    Sunhee Kim\textsuperscript{3},
    Minhwa Chung\textsuperscript{1}
}

\address{
    Department of Linguistics, Seoul National University, Republic of Korea\textsuperscript{1} \\
    Department of Computer Science and Engineering, Sogang University, Republic of Korea\textsuperscript{2} \\
    Department of French Language Education, Seoul National University, Republic of Korea\textsuperscript{3} \\
}

\begin{document}

\maketitle
\input{sections/0_abstract}

\input{sections/1_introduction}
\input{sections/2_method}
\input{sections/3_experiment}
\input{sections/4_analysis}
\input{sections/5_conclusion}

\section{Acknowledgment}
This work was supported by Institute of Information \& communications Technology Planning \& Evaluation (IITP) grant funded by the Korea government (MSIT) (No.2022-0-00223, Development of digital therapeutics to improve communication ability of autism spectrum disorder patients).

\bibliographystyle{IEEEbib}
\bibliography{refs}

\end{document}

%% file: sections/0_abstract.tex
\begin{abstract}
Automatic assessment of dysarthric speech is essential for sustained treatments and rehabilitation. However, obtaining atypical speech is challenging, often leading to data scarcity issues. To tackle the problem, we propose a novel automatic severity assessment method for dysarthric speech, using the self-supervised model in conjunction with multi-task learning. Wav2vec 2.0 XLS-R is jointly trained for two different tasks: severity classification and auxiliary automatic speech recognition (ASR). For the baseline experiments, we employ hand-crafted acoustic features and machine learning classifiers such as SVM, MLP, and XGBoost. 
Explored on the Korean dysarthric speech QoLT database, our model outperforms the traditional baseline methods, with a relative percentage increase of 1.25\% for F1-score. In addition, the proposed model surpasses the model trained without ASR head, achieving 10.61\% relative percentage improvements. Furthermore, we present how multi-task learning affects the severity classification performance by analyzing the latent representations and regularization effect.
\end{abstract}

\begin{keywords}
dysarthric speech, automatic assessment, self-supervised learning, multi-task learning
\end{keywords}

%% file: sections/1_introduction.tex
\section{Introduction} \label{sec:intro}
Dysarthria is a group of motor speech disorders resulting from neuromuscular control disturbances, which affects diverse speech dimensions such as respiration, phonation, resonance, articulation, and prosody \cite{darley1969differential}. Accordingly, people with dysarthria often suffer from degraded speech intelligibility, repeated communication failures, and, consequently, poor quality of life. 
Hence, accurate and reliable speech assessment is essential in the clinical field, as it helps track the condition of patients and the effectiveness of treatments.

The most common way of assessing severity levels of dysarthria is by conducting standardized tests such as Frenchay Dysarthria Assessment (FDA) \cite{enderby1980frenchay}. 
However, these tests heavily rely on human perceptual evaluations, which can be subjective and laborious. 
Therefore, automatic assessments that are highly consistent with the experts will have great potential for assisting clinicians in diagnosis and therapy.

Research on automatic assessment of dysarthria can be grouped into two approaches.
The first is to investigate a novel feature set.
For instance, paralinguistic features such as eGeMAPS were explored on their usability for atypical speech analysis \cite{xue2019acoustic}.
On the other hand, common symptoms of dysarthric speech provided insights into new feature sets - glottal \cite{narendra2020automatic}, resonance \cite{castillo2008automatic}, pronunciation \cite{novotny2014automatic, yeo2021automatic}, and prosody features \cite{kadi2013discriminative, hernandez2020prosody}.
Furthermore, representations extracted from deep neural networks were also examined, such as spectro-temporal subspace \cite{geng2022spectro}, i-vectors \cite{joshy2022automated}, and deepspeech posteriors \cite{tripathi2020improved}.
While this approach can provide intuitive descriptions of the acoustic cues used in assessments, it has the drawback of losing the information that may be valuable to the task.

The second approach is to explore the network architectures which take raw waveforms as input.
Networks include but are not limited to distance-based neural networks \cite{janbakhshi2021automatic}, LSTM-based models \cite{mayle2019diagnosing, bhat2020automatic} and CNN-RNN hybrid models \cite{shih2022dysarthria, ye2022hybrid}. 
As neural networks are often data-hungry, many researchers suffer from the data scarcity of atypical speech.
Consequently, research has often been limited to dysarthria detection, which is a binary classification task.
However, multi-class classification should also be considered for more detailed diagnoses.
Recently, self-supervised representation learning has arisen to alleviate such problems, presenting successes in various downstream tasks with a small amount of data \cite{baevski2020wav2vec,babu2021xls}.
Promising results were also reported for different tasks for atypical speech, including automatic speech recognition (ASR) \cite{hernandez2022cross,violeta2022investigating} and assessments \cite{jiang2021towards,bayerl2022detecting,getman2022wav2vec2}.
However, limited explorations were performed on the severity assessment of dysarthric speech.

This paper proposes a novel automatic severity classification method for dysarthric speech using a self-supervised learning model fine-tuned with multi-task learning (MTL). 
The model handles 1) a five-way multi-class classification of dysarthria severity levels as the main task and 2) automatic speech recognition as the auxiliary task. 
We expect MTL to have two advantages for the automatic severity classification of dysarthria.
First, the model is enforced to learn both acoustic and phonetic/pronunciation features for severity classification.
We hypothesize using these two complementary information can boost classification results.
Second, the auxiliary ASR task can act as a regularizer, as the model is trained to focus on two different tasks. 
This can prevent overfitting to small data and yield better classification performances.

The rest of the paper is organized as follows: \Cref{sec:method} introduces the proposed method, which consists of a self-supervised pre-trained model and fine-tuning with MTL.
\Cref{sec:experiment} describes the overall experimental settings and classification results.
Then, \Cref{sec:analysis} conducts further examinations, explaining how MTL can be so powerful.
Finally, \Cref{sec:con} is followed with a conclusion.



%% file: sections/2_method.tex
\input{figures/method}
\section{Method} \label{sec:method}
This section demonstrates our automatic severity classification method for dysarthric speech.
The overview of the proposed method can be found at \Cref{figs:method}.
First, we briefly introduce the self-supervised pre-trained wav2vec-based models.
Then, we describe the architectural modifications on the pre-trained model for multi-task learning. The model is fine-tuned on two tasks simultaneously: severity classification as the main task and ASR as the auxiliary task. 
We release the source codes of all the experiments for ease of reproduction\footnote{\url{https://github.com/juice500ml/dysarthria-mtl}}.

\subsection{Self-supervised pre-trained model} \label{ssec:pre-trained}
The key idea of self-supervised learning (SSL) models is to employ abundant unlabeled data to train a general speech model, namely, a self-supervised pre-trained model. 
Leveraged by the learned representations, the models have demonstrated promising results on downstream tasks by fine-tuning with the limited size of datasets \cite{baevski2020wav2vec}. 
We expect the dysarthric speech domain, which often suffers from data scarcity, can also take advantage of this approach.

\subsection{Fine-tuning by multi-task learning} \label{ssec:MTL}
We are motivated by the fact that speech intelligibility degrades with worse severity. 
Therefore, making decisions based on both acoustic and phonetic/pronunciation features may boost classification results.
To embody this domain knowledge into the severity classifier, we simultaneously trained the phoneme classifier as an auxiliary task.
Moreover, multi-task learning is considered as a regularization method that helps avoid overfitting.
In this study, the most simple variant of MTL is employed for the two classifiers: hard parameter sharing with a linear combination of losses.
Concretely, the two classifiers share the self-supervised pre-trained model, with separate linear heads for each task.

Firstly, the raw audio signal $\mathbf{x} \in [-1, 1]^L$ with length $L$ is fed into the model to yield $T$ latent speech representations $\mathbf{H} = [\mathbf{h}_1 \mathbf{h}_2 \dots \mathbf{h}_T] \in \mathbb{R}^{T \times F}$ of feature dimension $F$.

For the classification head, we average the latent representations, which are then passed through the fully connected layer to yield logits for five-way severity classification, following \cite{bayerl2022detecting}:
\begin{equation}
    \overline{\mathbf{h}} = \mathbb{E} [\mathbf{h}_t] = \textstyle\frac{1}{T} \textstyle\sum_{t=1}^T \mathbf{h}_t,
\end{equation}
\begin{equation}
    p_\texttt{CE}(y|\mathbf{x}) = \texttt{softmax}(\mathbf{W}_\texttt{CE}\overline{\mathbf{h}} + \mathbf{b}_\texttt{CE}),
\end{equation}
where $\overline{\mathbf{h}}$ is the averaged representation, $\mathbf{W}_\texttt{CE} \in \mathbb{R}^{5 \times F}$ and $\mathbf{b}_\texttt{CE} \in \mathbb{R}^{5 \times 1} $ is the learnable weights and biases of the fully connected layer.
Finally, we apply the cross-entropy loss $\mathcal{L}_\texttt{CE}$ on the classification predictions $p_\texttt{CE}$.

As for the ASR head, following \cite{baevski2020wav2vec,babu2021xls}, we pass each latent representation $\mathbf{h}_t$ at timestep $t$ through the common fully connected layer:
\begin{equation}
    p_\texttt{CTC}^t = \texttt{softmax}(\mathbf{W}_\texttt{CTC}\mathbf{h}_t + \mathbf{b}_\texttt{CTC}),
\end{equation}
where $p_\texttt{CTC}^t$ is the ASR prediction at timestep $t$, $\mathbf{W}_\texttt{CTC} \in \mathbb{R}^{V \times F}$ and $\mathbf{b}_\texttt{CTC} \in \mathbb{R}^{V \times 1} $ each refers to the learnable weights and biases of the fully connected layer, and $V$ is the size of the vocabulary.
We apply the Connectionist Temporal
Classification (CTC) loss $\mathcal{L}_\texttt{CTC}$ on stepwise ASR predictions $p_\texttt{CTC}^t$.

The final loss $\mathcal{L}$ is designed as the linear combination of two losses:
\begin{equation}
    \mathcal{L} = \mathcal{L}_\texttt{CE} + \alpha \mathcal{L}_\texttt{CTC},
\end{equation}
where $\alpha \in \mathbb{R}$ is the hyperparameter that balances the influence between two tasks.
As $\mathcal{L}_\texttt{CTC}$ tends to be few magnitudes larger than $\mathcal{L}_\texttt{CE}$, $\alpha \in (1.0, 0.1, 0.01, 0.001)$ are searched.

In addition, the convergence speed differs hugely between $\mathcal{L}_\texttt{CTC}$ and $\mathcal{L}_\texttt{CE}$.
Concretely, $\mathcal{L}_\texttt{CE}$ converges much quicker, hence overfits before $\mathcal{L}_\texttt{CTC}$ converges.
To mitigate the problem, we use only $\mathcal{L}_\texttt{CTC}$ in the initial $e$ epochs of training.
We test $e \in (0, 10, 20, 30, 40)$ out of $100$ epochs. 
Experiments regarding the effect of $\alpha$ and $e$ are demonstrated in \Cref{ssec:regularization}.

%% file: figures/method.tex
\begin{figure}[t] 
\centering
\includegraphics[width=0.97\columnwidth]{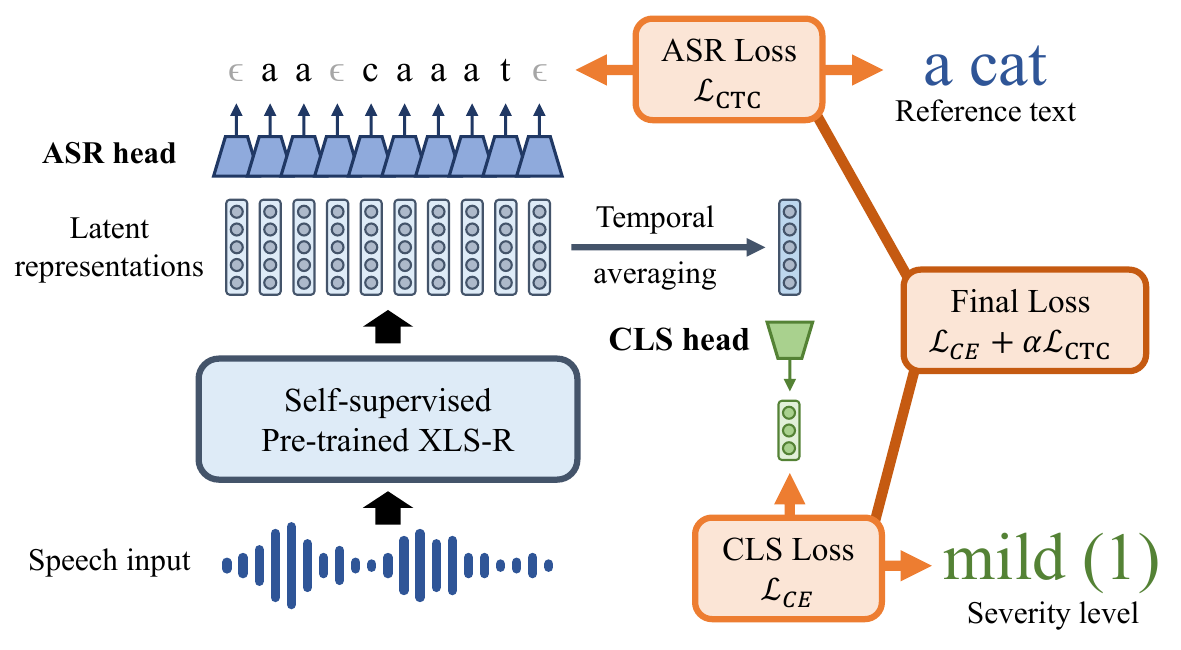}
\caption{
Illustration of our proposed method.
}
\label{figs:method}
\end{figure}

%% file: sections/3_experiment.tex
\section{Experiments} \label{sec:experiment}
\subsection{Dataset} \label{ssec:dataset}
Quality of Life Technology (QoLT) dataset \cite{choi2012dysarthric} is a Korean dysarthric speech corpus.
The corpus contains utterances from 10 healthy speakers (5 males, 5 females) and 70 dysarthric speakers (45 males, 25 females), where five speech pathologists conducted intelligibility assessments on a 5-point Likert scale: healthy (0), mild (1), mild-to-moderate (2), moderate-to-severe (3), and severe (4).
Accordingly, we use 25, 26, 12, 7 speakers for each severity level. 

QoLT dataset contains isolated words and restricted sentences, but only sentences are used for this study, similar to \cite{yeo2021automatic, hernandez2020prosody}.
Hence, a total of 800 utterances are used, consisting of five sentences recorded twice per speaker.
For the experiment, we conduct 5-way cross-validation in a speaker-independent manner.
The split is also stratified by gender. 


\vspace{-0.7em}
\subsection{Experimental details} \label{ssec:setup}
Following the findings from \cite{hernandez2022cross}, we choose XLS-R \cite{babu2021xls} with 300M parameters, which is the self-supervised model trained with cross-lingual data, as the pre-trained model. 
We optimize the batch size of $4$ using the Adam optimizer for $100$ epochs.
Similar to \cite{babu2021xls}, we use the Adam parameters with learning rate $\gamma = 2\times10^{-5}$, $\beta_1=0.9$, $\beta_2=0.98$, and $\epsilon=10^{-8}$.
Using the validation set, we keep the best model with the final loss $\mathcal{L}$.
We report the best classification accuracy achieved through grid-searching the hyperparameters, with the optimal values being $\alpha=0.1$ and $e=30$ in \Cref{tab:results}.

\subsection{Baselines} \label{ssec:baselines}
For baseline features, we use (1) eGeMAPS features, (2) hand-crafted features, and (3) their combination. 
First of all, we extract \textbf{eGeMAPS} feature set with the openSMILE toolkit \cite{eyben2010opensmile}. 
Consisting of 25 low-level descriptors (LLDs) with 88 features, eGeMaps is a basic standard acoustic parameter set designed to capture various aspects of speech, including frequency-, energy-, spectral-, and temporal-related features. 
We also extract \textbf{hand-crafted features} used from the previous studies \cite{yeo2021automatic, yeo2022cross}.
The feature list was proposed to capture the common symptoms of dysarthria at different speech dimensions, such as voice quality, pronunciation (phoneme correctness, vowel distortion), and prosody (speech rate, pitch, loudness, rhythm).
For the \textbf{combined feature set}, we simply concatenate the two feature lists.

Regarding classification, we apply three classifiers that showed successful results for dysarthria severity classification \cite{yeo2021automatic, hernandez2020prosody, yeo2022cross}: support vector machine (SVM), multi-layer perceptron (MLP), and XGBoost. 
The hyperparameters of the classifiers are optimized using grid search on the validation set.
We optimize SVM with a radial basis kernel function in terms of $C$ and $\gamma$, by grid-searching both parameters between $10^{-4}$ and $10^4$.
As for MLP, the best number of hidden layers, activation function, optimizer, and learning rate are searched. 
The number of hidden layers is checked between 1 and 10, activation function among tanh, relu, logistic, identity function, optimizer between Adam and SGD, and learning rate between $10^{-4}$ and $10^{-1}$.
For XGBoost, we tested maximum depth between $3$ and $5$.
To validate the effectiveness of multi-task learning (proposed; MTL), we also conduct a single-task learning (\textbf{STL}) experiment, where the self-supervised model is fine-tuned without an auxiliary ASR task: hyperparameter settings are the same as MTL, only different in setting $\alpha$ as $0$, to use only the classification loss $\mathcal{L}_\texttt{CE}$ for training.

\subsection{Results} \label{ssec:result}

\Cref{tab:results} presents the performance of traditional baselines with fine-tuned SSL models, by using classification accuracy, precision (macro), recall (macro), and F1-score (macro).
The best performance for each metric is indicated in bold. 

First, comparisons within the traditional baseline experiments demonstrate that using the hand-crafted feature set achieves better performances than using eGeMAPS for all classifiers. 
When the eGeMAPS and hand-crafted features are combined, SVM and MLP give worse performances than using hand-crafted features.
This may be due to a large feature vector, which causes overfitting and reduced performance of the classifiers.
On the other hand, XGBoost slightly outperforms the hand-crafted features, where the algorithm is less prone to overfitting.
However, the performance does not reach the SVM with hand-crafted features.

\input{figures/results} 

Second, we compare the performances of the SSL model with single-task learning (STL) and the traditional baselines. 
With 61.02\% classification accuracy and 57.13\% F1-score, STL shows better performances than all baselines except for the experiment using the hand-crafted feature set for SVM classifier, which attains 61.02\% classification accuracy and 62.41\% F1-score.

Lastly, we analyze the performance of our proposed method (MTL) in contrast to the baseline experiments. Experimental results demonstrates that MTL achieves the highest accuracy, precision, recall, and F1-score, with 65.52\%, 66.47\%, 64.86\%, 63.19\%, respectively. 
This is the relative increase of 7.37\%, 3.55\%, 2.64\%, 1.25\% compared to the best-performing baseline, SVM using the hand-crafted feature set.

%% file: figures/results.tex
\begin{table}[t]
\caption{
Classification performance compared to the baselines.
}
\label{tab:results}
\centering
\resizebox{0.45\textwidth}{!}
{

\begin{tabular}{c|c|c|c|c|c}
\hline
Input & Classifier & Accuracy & Precision & Recall & F1-score\\
\hline
\multirow{3}{*}{eGeMAPS}
& SVM & 55.01 & 53.89 & 53.27 & 52.28 \\
& MLP & 50.79 & 44.46 & 48.60 & 46.58 \\
& XGBoost & 52.20 & 55.07 & 50.85 & 50.61 \\
\hline
\multirow{3}{*}{\shortstack{Hand-crafted\\ features}}
& SVM & 61.02 & 64.19 & 63.19 & 62.41 \\
& MLP & 55.74 & 60.06 & 60.34  & 58.85 \\
& XGBoost & 55.72 & 61.14 & 56.21 & 56.16 \\
\hline
\multirow{3}{*}{\shortstack{eGeMaps\\ + Hand-crafted \\ features}}
& SVM & 57.83 & 58.83 & 57.59 & 56.65 \\
& MLP & 50.21 & 48.40 & 47.31 & 46.76 \\
& XGBoost & 56.29 & 62.29 & 56.23 & 56.68 \\
\hline
\multirow{2}{*}{Raw audio} 
& STL & 61.02 & 64.09 & 57.93 & 57.13 \\
& \textbf{MTL} & \textbf{65.52} & \textbf{66.47} & \textbf{64.86} & \textbf{63.19} \\
\hline
\end{tabular}

}
\vspace{-1em}
\end{table}

%

%% file: sections/4_analysis.tex
\section{Anaylsis} \label{sec:analysis}
\subsection{Analysis on latent representations} \label{ssec:representations}
\Cref{figs:representation} visualizes the latent representations of the training set samples to observe how fine-tuning shapes the latent representation space.
For the analysis, we use the fully-converged model instead of the best-kept model, to demonstrate the representation space learned by the loss.
Averaged representations $\overline{\mathbf{h}}$ are visualized by using UMAP \cite{mcinnes2018umap}.

As presented in \Cref{figs:representation}, the representations from the STL model cannot be distinguished by different sentences, while the representations are clustered in terms of both sentences and severity levels for the MTL model.
The analysis indicates that the MTL model also encodes phonetic/pronunciation information.
Note that unlike other severity levels, different utterances from severe (4) dysarthric speakers are strongly clustered.
We assume this may be due to significantly distorted speech, which makes it difficult for the ASR head to separate their representations.
\input{figures/representation}

\subsection{Analysis on the regularization effect} \label{ssec:regularization}
\Cref{fig:regularization} presents the effect of MTL over STL and the efficacy of postponing the $\mathcal{L}_\texttt{CE}$ optimization with the hyperparameter $e$.
With joint optimization of CTC loss $\mathcal{L}_\texttt{CTC}$ and CE loss $\mathcal{L}_\texttt{CE}$, $\mathcal{L}_\texttt{CE}$ overfits much slower than STL, which implies MTL's regularization effect.
Moreover, stable optimization and better performances are found on both classification and ASR tasks with $e$ set to nonzero.

\Cref{tab:ablation} further shows the effect the hyperparameters $\alpha$ and $e$.
Bigger the $\alpha$ and later the $e$, the phone error rate (PER) consistently drops, since the emphasis on the $\mathcal{L}_\texttt{CTC}$ increases.
With the best accuracy found in the mid-point of the hyperparameter grid, 
this validates the effectiveness of aligning the convergence speed of the two losses.
We suspect that premature training of $\mathcal{L}_\texttt{CE}$ leads to the model being under-trained with the ASR task, which fails to inject enough information, resulting in worse classification performance.

\input{figures/regularization}
\input{figures/ablation}




%% file: figures/representation.tex
\begin{figure}[t] 
\centering
\vspace{-2em}
\subfloat[Trained with MTL,\\color-coded with severity.]{\includegraphics[width=0.43\columnwidth]{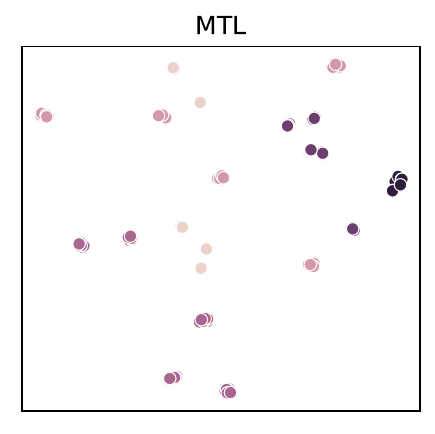}}
\subfloat[Trained with STL,\\color-coded with severity.]{\includegraphics[width=0.43\columnwidth]{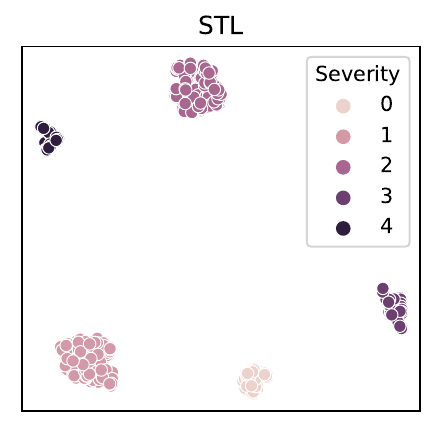}}\\
\vspace{-1em}
\subfloat[Trained with MTL,\\color-coded with setence.]{\includegraphics[width=0.43\columnwidth]{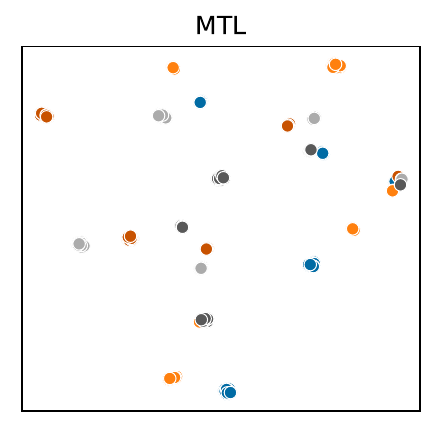}}
\subfloat[Trained with STL,\\color-coded with setence.]{\includegraphics[width=0.43\columnwidth]{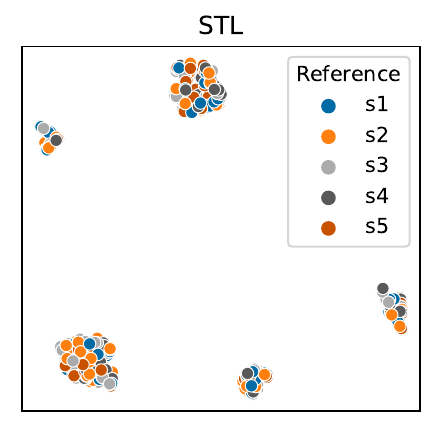}}
\caption{
Averaged representation $\overline{\mathbf{h}}$ of the training set samples.
$0$ to $4$ refers to severity levels, starting from healthy to severe.
$s$ in $s1...s2$ refers to sentences.
}
\vspace{-1em}
\label{figs:representation}
\end{figure}

%% file: figures/regularization.tex
\begin{figure}[t!] 
\centering
\vspace{-1.5em}
\includegraphics[width=0.9\columnwidth]{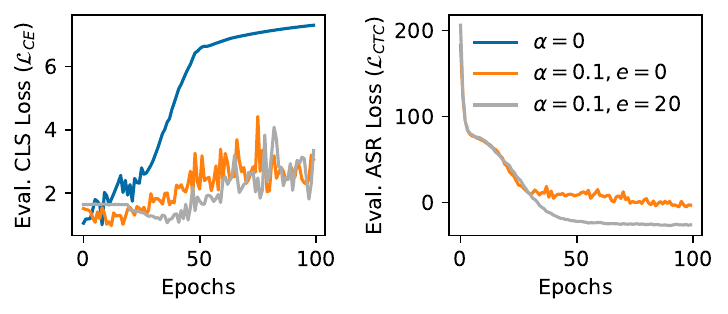}
\caption{
Classification loss $\mathcal{L}_\texttt{CE}$ and ASR loss $\mathcal{L}_\texttt{CTC}$ on validation set.
$\alpha=0$ denotes the STL case when we use the $\mathcal{L}_\texttt{CE}$ only.
}
\label{fig:regularization}
\end{figure}

%% file: figures/ablation.tex
\begin{table}[t]
\caption{
Ablation study on hyperparameters $\alpha$ and $e$.
}
\label{tab:ablation}
\centering
\resizebox{0.4\textwidth}{!}
{
\begin{tabular}{c|c|c|c|c}

\hline
Accuracy & $\alpha=1.0$ & $\alpha=0.1$ & $\alpha=0.01$ & $\alpha=0.001$ \\
\hline
$e=0$  & 60.51 & 60.69 & 56.21 & 54.94 \\
\hline
$e=10$ & 61.82 & 63.12 & 57.14 & 57.00 \\
\hline
$e=20$ & 54.77 & 64.69 & 59.84 & 61.27 \\
\hline
$e=30$ & 57.74 & \textbf{65.52} & 60.10 & 62.72 \\
\hline
$e=40$ & 55.47 & 60.11 & 62.00 & 57.96 \\
\hline
\hline
PER & $\alpha=1.0$ & $\alpha=0.1$ & $\alpha=0.01$ & $\alpha=0.001$ \\
\hline
$e=0$  & 17.50 & 21.86 & 88.49 & 96.91 \\
\hline
$e=10$ & 14.83 & 22.37 & 82.59 & 96.49 \\
\hline
$e=20$ & 16.66 & 18.10 & 31.12 & 90.08 \\
\hline
$e=30$ & 15.87 & 17.72 & 23.10 & 74.54 \\
\hline
$e=40$ & 15.41 & 15.95 & 20.45 & 56.24 \\
\hline
\end{tabular}
}

\vspace{-1em}

\end{table}

%

%% file: sections/5_conclusion.tex
\vspace{-0.5em}\section{Conclusion} \label{sec:con}
This paper proposes a novel automatic dysarthria severity classification method: a self-supervised model fine-tuned with multi-task learning, jointly learning the five-way severity classification task and the ASR task.
Our proposed model outperforms the traditional baseline experiments, which employ eGeMaps, hand-crafted feature sets as input, and SVM, MLP, and XGBoost as the classifier.
Further analyses regarding the latent representations and regularization effect provide explanations for how our proposed MTL method could be effective.
For future work, we plan to investigate more complex MTL settings for further performance improvements.
Extending the applications of the proposed method to different corpora and languages is also necessary.






%% file: main.bbl
\begin{thebibliography}{10}

\bibitem{darley1969differential}
FL~Darley, AE~Aronson, and JR~Brown,
\newblock ``Differential diagnostic patterns of dysarthria,''
\newblock {\em Journal of speech and hearing research}, 1969.

\bibitem{enderby1980frenchay}
P~Enderby,
\newblock ``Frenchay dysarthria assessment,''
\newblock {\em British Journal of Disorders of Communication}, 1980.

\bibitem{xue2019acoustic}
W~Xue, C~Cucchiarini, R~Hout, and H~Strik,
\newblock ``Acoustic correlates of speech intelligibility: the usability of the
  egemaps feature set for atypical speech,''
\newblock in {\em SLaTE}, 2019.

\bibitem{narendra2020automatic}
NP~Narendra and P~Alku,
\newblock ``Automatic intelligibility assessment of dysarthric speech using
  glottal parameters,''
\newblock {\em Speech Communication}, 2020.

\bibitem{castillo2008automatic}
E~Castillo-Guerra and W~Lee,
\newblock ``Automatic assessment of perturbations produced by audible
  inspirations in pathological voices,''
\newblock in {\em ASA}, 2008.

\bibitem{novotny2014automatic}
M~Novotn{\`y}, J~Rusz, R~{\v{C}}mejla, and E~R\r{u}{\v{z}}i{\v{c}}ka,
\newblock ``Automatic evaluation of articulatory disorders in parkinson’s
  disease,''
\newblock {\em TASLP}, 2014.

\bibitem{yeo2021automatic}
EJ~Yeo, S~Kim, and M~Chung,
\newblock ``Automatic severity classification of korean dysarthric speech using
  phoneme-level pronunciation features.,''
\newblock in {\em Interspeech}, 2021.

\bibitem{kadi2013discriminative}
KL~Kadi, SA~Selouani, B~Boudraa, and M~Boudraa,
\newblock ``Discriminative prosodic features to assess the dysarthria severity
  levels,''
\newblock in {\em World Congress on Engineering}, 2013.

\bibitem{hernandez2020prosody}
A~Hernandez, S~Kim, and M~Chung,
\newblock ``Prosody-based measures for automatic severity assessment of
  dysarthric speech,''
\newblock {\em Applied Sciences}, 2020.

\bibitem{geng2022spectro}
M~Geng, S~Liu, J~Yu, X~Xie, S~Hu, and et~al.,
\newblock ``Spectro-temporal deep features for disordered speech assessment and
  recognition,''
\newblock in {\em Interspeech}, 2021.

\bibitem{joshy2022automated}
AA~Joshy and R~Rajan,
\newblock ``Automated dysarthria severity classification: A study on acoustic
  features and deep learning techniques,''
\newblock {\em IEEE Transactions on Neural Systems and Rehabilitation
  Engineering}, 2022.

\bibitem{tripathi2020improved}
A~Tripathi, S~Bhosale, and SK~Kopparapu,
\newblock ``Improved speaker independent dysarthria intelligibility
  classification using deepspeech posteriors,''
\newblock in {\em ICASSP}, 2020.

\bibitem{janbakhshi2021automatic}
P~Janbakhshi, I~Kodrasi, and H~Bourlard,
\newblock ``Automatic dysarthric speech detection exploiting pairwise
  distance-based convolutional neural networks,''
\newblock in {\em ICASSP}, 2021.

\bibitem{mayle2019diagnosing}
A~Mayle, Z~Mou, R~C Bunescu, S~Mirshekarian, L~Xu, and C~Liu,
\newblock ``Diagnosing dysarthria with long short-term memory networks.,''
\newblock in {\em Interspeech}, 2019.

\bibitem{bhat2020automatic}
C~Bhat and H~Strik,
\newblock ``Automatic assessment of sentence-level dysarthria intelligibility
  using blstm,''
\newblock {\em IEEE J. Sel. Top. Signal Process}, 2020.

\bibitem{shih2022dysarthria}
D~Shih, C~Liao, T~Wu, X~Xu, and M~Shih,
\newblock ``Dysarthria speech detection using convolutional neural networks
  with gated recurrent unit,''
\newblock in {\em Healthcare}, 2022.

\bibitem{ye2022hybrid}
W~Ye, Z~Jiang, Q~Li, Y~Liu, and Z~Mou,
\newblock ``A hybrid model for pathological voice recognition of post-stroke
  dysarthria by using 1dcnn and double-lstm networks,''
\newblock {\em Applied Acoustics}, 2022.

\bibitem{baevski2020wav2vec}
A~Baevski, Y~Zhou, A~Mohamed, and M~Auli,
\newblock ``wav2vec 2.0: A framework for self-supervised learning of speech
  representations,''
\newblock {\em NeurIPS}, 2020.

\bibitem{babu2021xls}
A~Babu, C~Wang, A~Tjandra, K~Lakhotia, Q~Xu, et~al.,
\newblock ``{XLS-R:} self-supervised cross-lingual speech representation
  learning at scale,''
\newblock in {\em Interspeech}, 2022.

\bibitem{hernandez2022cross}
A~Hernandez, PA~P{\'e}rez-Toro, E~N{\"o}th, JR~Orozco-Arroyave, A~Maier, and
  SH~Yang,
\newblock ``Cross-lingual self-supervised speech representations for improved
  dysarthric speech recognition,''
\newblock in {\em Interspeech}, 2022.

\bibitem{violeta2022investigating}
LP~Violeta, WC~Huang, and T~Toda,
\newblock ``Investigating self-supervised pretraining frameworks for
  pathological speech recognition,''
\newblock in {\em Interspeech}, 2022.

\bibitem{jiang2021towards}
H~Jiang, WYB Lim, JS~Ng, Y~Wang, Y~Chi, and C~Miao,
\newblock ``Towards parkinson’s disease prognosis using self-supervised
  learning and anomaly detection,''
\newblock in {\em ICASSP}, 2021.

\bibitem{bayerl2022detecting}
Sebastian~P B, Dominik W, Elmar N, and Korbinian R,
\newblock ``Detecting dysfluencies in stuttering therapy using wav2vec 2.0,''
\newblock in {\em Interspeech}, 2022.

\bibitem{getman2022wav2vec2}
Y~Getman, R~Al-Ghezi, E~Voskoboinik, T~Gr{\'o}sz, M~Kurimo, G~Salvi,
  T~Svendsen, and S~Str{\"o}mbergsson,
\newblock ``wav2vec2-based speech rating system for children with speech sound
  disorder,''
\newblock in {\em Interspeech}, 2022.

\bibitem{choi2012dysarthric}
D~Choi, B~Kim, Y~Kim, Y~Lee, Y~Um, and M~Chung,
\newblock ``Dysarthric speech database for development of qolt software
  technology,''
\newblock in {\em LREC}, 2012.

\bibitem{eyben2010opensmile}
F~Eyben, M~W{\"o}llmer, and B~Schuller,
\newblock ``Opensmile: the munich versatile and fast open-source audio feature
  extractor,''
\newblock in {\em ACM on Multimedia}, 2010.

\bibitem{yeo2022cross}
E~J Yeo, K~Choi, S~Kim, and M~Chung,
\newblock ``Cross-lingual dysarthria severity classification for english,
  korean, and tamil,''
\newblock {\em arXiv preprint arXiv:2209.12942}, 2022.

\bibitem{mcinnes2018umap}
L~McInnes, J~Healy, and J~Melville,
\newblock ``Umap: Uniform manifold approximation and projection for dimension
  reduction,''
\newblock {\em arXiv preprint arXiv:1802.03426}, 2018.

\end{thebibliography}
